\documentclass[conference]{IEEEtran}
\IEEEoverridecommandlockouts

\usepackage{cite}
\usepackage{amsmath,amssymb,amsfonts}
\usepackage{algorithmic}
\usepackage{graphicx}
\usepackage{textcomp}
\usepackage{xcolor}
\usepackage{hyperref}
\usepackage{algorithm}
\usepackage[normalem]{ulem}
\useunder{\uline}{\ul}{}
\usepackage{subcaption}
\usepackage{multirow}
\usepackage{caption}
\usepackage{comment}

\def\BibTeX{{\rm B\kern-.05em{\sc i\kern-.025em b}\kern-.08em
    T\kern-.1667em\lower.7ex\hbox{E}\kern-.125emX}}
\begin{document}

\title{No Labels Needed: Zero-Shot Image Classification with Collaborative Self-Learning\\
\thanks{This paper was accepted at International Conference on Tools with Artificial Intelligence (ICTAI) 2025 and the work was supported by the Coordination for the Improvement of Higher Education Personnel (CAPES) - Finance code 001; and by the Brazilian National Council for Scientific and Technological Development (CNPq).}
}


\author{\IEEEauthorblockN{1\textsuperscript{st} Matheus Vinícius Todescato}
\IEEEauthorblockA{\textit{Institute of Informatics} \\
\textit{UFRGS}\\
Porto Alegre, Brazil \\
mvtodescato@inf.ufrgs.br}
\and
\IEEEauthorblockN{2\textsuperscript{nd} Joel Luís Carbonera}
\IEEEauthorblockA{\textit{Institute of Informatics} \\
\textit{UFRGS}\\
Porto Alegre, Brazil \\
jlcarbonera@inf.ufrgs.br}
}

\maketitle

\begin{abstract}
While deep learning, including Convolutional Neural Networks (CNNs) and Vision Transformers (ViTs), has significantly advanced classification performance, its typical reliance on extensive annotated datasets presents a major obstacle in many practical scenarios where such data is scarce. Vision-language models (VLMs) and transfer learning with pre-trained visual models appear as promising techniques to deal with this problem. This paper proposes a novel zero-shot image classification framework that combines a VLM and a pre-trained visual model within a self-learning cycle. Requiring only the set of class names and no labeled training data, our method utilizes a confidence-based pseudo-labeling strategy to train a lightweight classifier directly on the test data, enabling dynamic adaptation. The VLM identifies high-confidence samples, and the pre-trained visual model enhances their visual representations. These enhanced features then iteratively train the classifier, allowing the system to capture complementary semantic and visual cues without supervision. Notably, our approach avoids VLM fine-tuning and the use of large language models, relying on the visual-only model to reduce the dependence on semantic representation. Experimental evaluations on ten diverse datasets demonstrate that our approach outperforms the baseline zero-shot method.
\end{abstract}


\section{Introduction}


Recent advancements in deep learning have significantly improved image classification performance, making these techniques dominant in current research. Deep learning-based classification methods are now employed in diverse areas such as Biomedicine \cite{alzubaidi:21}, Healthcare \cite{abbas2021classification}, and Geology \cite{todescato2024multiscale}, etc. In general, complex deep learning models require extensively annotated datasets to perform effectively \cite{zhu2021investigation}. However, annotated data is scarce in many practical applications, creating a significant obstacle to successfully deploying deep learning models. 

In this context, vision-language models (VLMs) \cite{radford2021learning} have shown strong potential for classification tasks, achieving high performance without requiring additional labeled data. These models are trained on large datasets of image-text pairs from the internet, performing a process of contrastive learning, learning to pair images with their correct captions. In addition to this type of strategy, in order to have better performance in downstream image classification tasks, several works take advantage of transfer learning using visual models pre-trained on large image datasets \cite{torrey2010transfer} to extract informative features from the data without the need for specific training \cite{zhuang:20}. With these features, it is possible to perform training in a simple image classifier  \cite{mallouh:19}, achieving high performance in the classification task \cite{todescato2024investigating,iceis24erick}.

Despite recent progress in zero-shot image classification using vision-language models (VLMs), current approaches often rely on prompt engineering, large language models (LLMs), or fine-tuning strategies that can be computationally expensive and domain-sensitive. These dependencies limit the applicability of existing methods in settings where labeled data is scarce, class semantics are subtle, or infrastructure constraints prevent large-scale fine-tuning. Moreover, most methods tightly couple semantic and visual spaces, which may propagate bias and hinder generalization.

This paper proposes a novel zero-shot image classification framework that combines in an original way a vision-language model and a pre-trained visual model within a self-learning cycle. Our method requires only the set of class names corresponding to the dataset's classes, eliminating the need for any labeled training data. The framework utilizes a confidence-based pseudo-labeling strategy to train a lightweight classifier directly on the test data, enabling dynamic adaptation to the target distribution.

Another key innovation of our approach is its modular design, which facilitates collaborative interaction between different vision-language models and pre-trained visual models. Specifically, we use the vision-language model to select the most confident samples during the self-learning cycle, while the pre-trained visual model is employed as a feature extractor to provide significant representations of the selected samples. These extracted features are then used to train a lightweight classifier. This iterative process allows the system to capture complementary semantic and visual cues without supervision. Experimental results show that our approach outperforms the baseline zero-shot method in image classification accuracy by a considerable margin.

The structure of this paper is as follows. Section \ref{sec:rel_works} reviews related works. Section \ref{sec:approach} presents a detailed description of the proposed approach. Section \ref{sec:experiments} details our experiments, encompassing datasets, methodology, and results. Lastly, Section \ref{sec:conclusion} summarizes the conclusions.

\section{Related Works}\label{sec:rel_works}

Transformer-based architectures \cite{dosovitskiy:20} have paved the way for sophisticated vision-language models (VLMs), revolutionizing the field of computer vision \cite{iceis24erick}. These models are pre-trained on massive datasets \cite{deng2009imagenet}, and their strong performance has led researchers to investigate their capabilities in zero-shot tasks.

Several VLMs leverage contrastive learning to achieve impressive zero-shot results. For example, FLAVA \cite{singh2022flava} applies learning from both paired and unpaired image and text data, using different loss functions to understand both multimodal and unimodal information. Similarly, ALIGN \cite{jia2021scaling} utilizes a vast amount of noisy image-text data, often sourced from image alt-texts, and trains using a contrastive loss to align visual and textual representations. CLIP (Contrastive Language-Image Pre-Training) \cite{radford2021learning} stands out among this model type. To combine text and images, CLIP trains an image encoder and a text encoder using image-text pairs, applying a similar objective function. To achieve this, during training, the function maximizes the similarity between the embedding of the image and the embedding of the text for each pair. The model has been trained on various images and textual labels that are abundantly available on the internet (more than 400 million pairs of images and text). Then, to use the model for testing, the similarity between the image embedding and the class label embeddings, such as “a photo of a [\textit{class name}]" is calculated, ranking the labels based on their similarity to the image, being able to perform top-1 or top-$x$ prediction. CLIP's capability to encode images and text into a shared latent space allows the comparison of a set of class names and images, promoting excellent results in zero-shot image classification and being the baseline for the task.

Most works try to improve CLIP results in this task by generating better prompts or fine-tuning the model. Several works perform prompt tuning by improving the textual description of classes \cite{gan2023decorate}. Some, like CoOp \cite{zhou2022learning}, append learnable context vectors to the class names. Other works, like CHiLS \cite{novack2023chils} and CuPL \cite{pratt2023does}, use improved class descriptions from GPT text generation \cite{brown2020language}. However, these methods do not adapt well to fine-grained domains (i.e., focusing on subcategories within a broader category) or highly specific domains, which are common in data-scarce environments, thereby limiting their applicability in such contexts \cite{saha2024improved}.

Among fine-tuning–based approaches, some works improve few-shot classification using few samples to parameter updates in CLIP \cite{gao2024clip}, depending on labeled data. For zero-shot, an example is LaFTer \cite{mirza2023lafter}, which uses unpaired images and texts obtained by querying LLMs (Large Language Models) to fine-tune the CLIP encoders. Other works apply the use of GPT to perform this fine-tuning \cite{lewis2023gist,naeem2023i2mvformer}. 
In \cite{saha2024improved}, the authors present three distinct approaches using attributes from different strategies. The first method augments the standard CLIP prompt with domain information. In the second approach, instead of using only the class name, they perform robust generation of descriptions using LLMs, which are concatenated to the standard prompt and include visual characteristics, habitat information, and other distinguishing features. In addition to these two, they also perform a fine-tuning of CLIP using these descriptions. The first two approaches are completely zero-shot, with the final approach being zero-shot only at inference, as it is trained on samples of \textit{seen classes} to generalize to \textit{unseen classes}. The results show that this work is superior to the baseline, being the state-of-the-art for the task. 
However, a key limitation of these methods is their reliance on large language models, which further increases dependence on the VLM joint embedding space, as they use information generated by the LLMs to expand the prompts or to fine-tune the vision-language model.

The key distinction between our work and previous approaches lies in using the default CLIP model without modifying its architecture or class descriptions. Instead of fine-tuning, our method leverages transfer learning by utilizing a pre-trained visual-only model, allowing us to extract representative visual features from images. This strategy has proven highly effective across various domains—particularly in medical applications such as cancer detection, pneumonia identification, and COVID-19 diagnosis—where transfer learning plays a crucial role in achieving strong performance \cite{kim2022transfer}.

The traditional self-learning methods refine the model using predictions on in-distribution data of the same model \cite{xie2020self}, being prone to \emph{confirmation bias} — where initial errors are reinforced through iterative training. To mitigate this, we decouple the sources of information: the VLM (CLIP, for example) is used to guide sample selection, while the pre-trained visual model (playing the role of feature extractor) provides independent features of the images. Building on this foundation, our pipeline promotes a collaborative self-learning framework, training a lightweight classifier that achieves strong performance when provided with informative features while remaining efficient to train \cite{todescato2024investigating,iceis24erick}. The idea behind this collaboration is to have a rich source of visual information for the classifier, filtering the information coming from CLIP, consequently reducing the dependence on the VLM's joint embedding space, thereby mitigating biases that may be present in its learned semantic representations. This dual-source strategy enables training a lightweight classifier without requiring labeled data. This combination to create a self-learning cycle is novel in the literature and can be applied to other tasks or with other backbones. It is powerful because it can adapt to the specific characteristics of the dataset in use.

\section{Proposed Approach} \label{sec:approach}

Given the good performance that can be achieved in conventional supervised image classification using only a classifier trained with features extracted by pre-trained models, our goal is to approximate this classification potential without using labeled data. To address this challenge, we propose a strategy to train a lightweight classifier on pseudo-labeled samples and refine it iteratively in a self-learning loop using features extracted from a pre-trained image model.

The proposed pipeline integrates three key components:
\begin{itemize}
\item Image/Text Encoder: A pre-trained model that embeds images and texts into a shared latent space to compute similarity scores. In this work, we use CLIP \cite{radford2021learning} to play this role. We refer to CLIP\textsubscript{text} as the encoder of texts and CLIP\textsubscript{image} as the encoder of images.
\item Feature Extractor: A pre-trained model (e.g., ViT-G-14) that transforms images into representative feature vectors.
\item Image Classifier: A lightweight model, iteratively trained on pseudo-labels in a self-learning loop, and whose logits are used to compute OOD scores of test samples.
\end{itemize}

These three components function together in a collaborative way to make the approach effective: CLIP guides the sample selection process, the pre-trained visual model supplies independent image features (rather than relying solely on CLIP's features), and together they enable the training of the image classifier.

\begin{figure*}[!ht]
\centering
\includegraphics[width=0.8\textwidth]{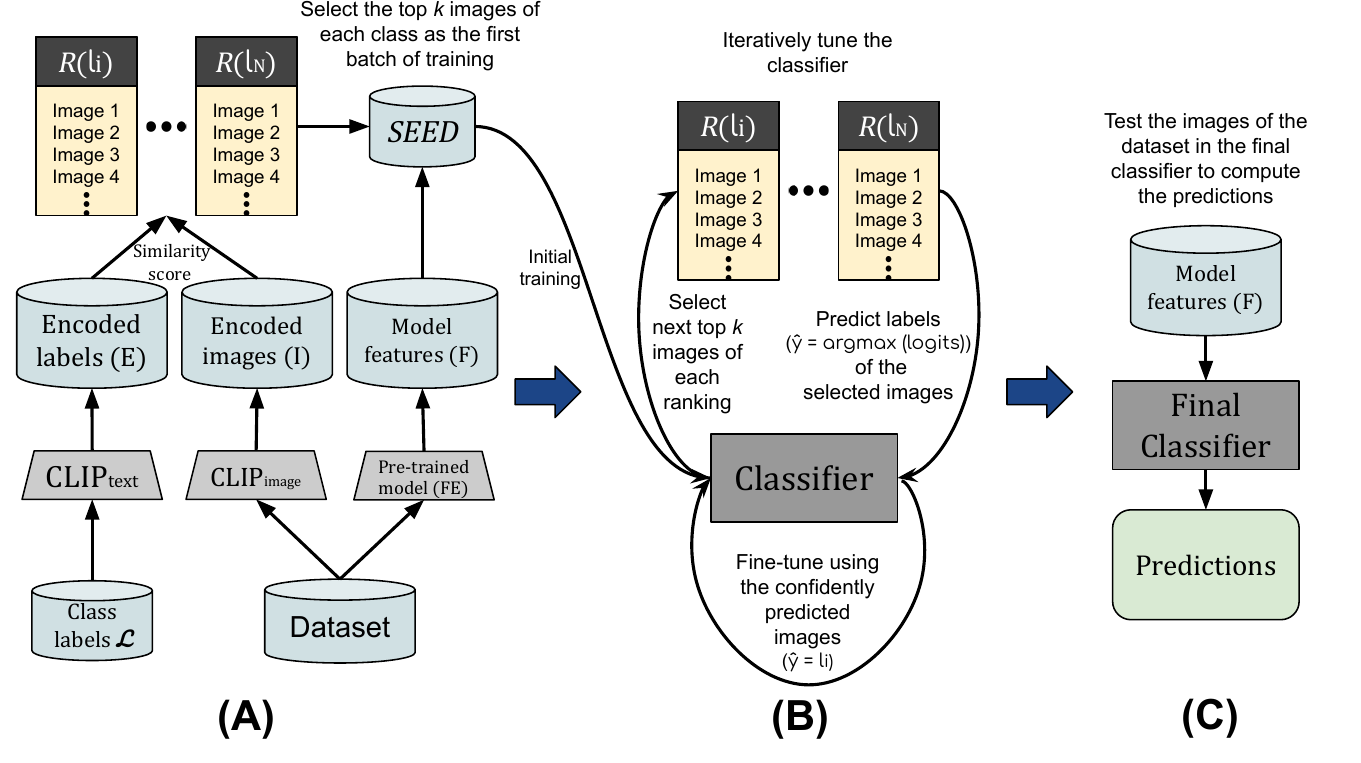}
\caption{Our approach pipeline is divided into three steps. In (A), we construct the initial training set ($SEED$) using CLIP and extract features from the images using a pre-trained model. Next (B), we perform a self-learning cycle, incrementally tuning a classifier. In the third step (C), we make the predictions for each image in the dataset using the trained classifier.}
\label{fig:diagram}
\end{figure*}

As we show in Figure \ref{fig:diagram} and Algorithm~\ref{alg:zeroshot}, our approach takes as input only a set of textual labels \( \mathcal{L} = \{l_1, l_2, \dots, l_N\} \) that represents the set of $N$ classes and an unlabeled dataset of images and proceeds through three sequential steps:
\begin{enumerate}
    \item \textbf{Step A) Seed Selection}: An initial set of training images is identified with high confidence of its true label.
    \item \textbf{Step B) Classifier Training}: A self-learning loop iteratively refines the classifier using pseudo-labels.
    \item \textbf{Step C) Image classification}: The trained classifier is used to predict and classify the images.
\end{enumerate}

In the following, we describe each step in detail.

\begin{algorithm}
\scriptsize
\caption{Zero-shot image classification with collaborative self-learning}
\label{alg:zeroshot}
\begin{algorithmic}
\REQUIRE Set of labels \( \mathcal{L} = \{l_1, l_2, \dots, l_N\} \), dataset of images, pre-trained feature extractor \( FE \), hyperparameters \( k \), \( I_{\text{epochs}} \), \( R_{\text{epochs}} \).
\ENSURE Class prediction for each image in the dataset

\STATE \textbf{Step A: Seed Selection}
\STATE Initialize \( SEED \leftarrow \emptyset \)
\STATE Encode all images in the dataset using CLIP\textsubscript{image} to generate image embeddings \( \{I_1, I_2, \dots, I_M\} \)
\STATE Extract features for all images using \( FE(x) \), resulting in feature vectors \( \{F_1, F_2, \dots, F_M\} \)
\FOR{each class label \( l_i \) in \( \mathcal{L} \)}
    \STATE Encode \( l_i \) using CLIP\textsubscript{text} to generate a text embedding \( E_i \)
    \FOR{each image embedding \( I_j \)}
        \STATE Compute cosine similarity (\( \text{cos}(I_j, E_i) \))
    \ENDFOR
    \STATE Select the most similar images to \( E_i \) and define set \( \mathcal{B}(l_i) \)
    
    \FOR{each image \( x \in \mathcal{B}(l_i) \)}
        \STATE Retrieve the \( k \) most similar images \( r_j(x) \), for \( j = 1, \dots, 5 \), based on cosine similarity to \( x \)'s embedding
        \FOR{each \( r_j(x) \)}
            \STATE Predict the most similar labels using CLIP
        \ENDFOR
        \STATE Compute score:
        \[
        S(x) = \frac{1}{k} \sum_{j=1}^{k} \cos(\mathcal{I}(r_j(x)), \mathcal{T}(l_i))
        \]
    \ENDFOR

    \STATE Construct a ranking \( \mathcal{R}(l_i) \) by sorting the images in \( \mathcal{B}(l_i) \) in descending order of \( S(x) \)
    \STATE Select top \( k \) images from \( \mathcal{R}(l_i) \)
     \STATE Assign \( l_i \) as pseudo-label to selected images and add feature vectors \( F_j \) to \( SEED \)
\ENDFOR

\STATE \textbf{Step B: Classifier Training} 
\STATE Train an initial classifier using \( SEED \) for \( I_{\text{epochs}} \)
\REPEAT
    \FOR{each label \( l_i \in \mathcal{L} \)}
        \STATE Select next top \( k \) images from \( \mathcal{R}(l_i) \)
        \STATE Using the classifier generate a predicted label $\hat{y} = \arg\max(\text{logits})$ for each image.
        \STATE Assign the \( l_i \) label to confidently predicted images ($\hat{y} = l_i$) and add their feature vectors \( F_j \) to \(\mathcal{D}_{\text{tune}}\)
    \ENDFOR
    \STATE Fine-tune the classifier for \( R_{\text{epochs}} \) using \(\mathcal{D}_{\text{tune}}\)
    \STATE Reset the set \(D_{\text{tune}} \leftarrow \emptyset\)
\UNTIL{stopping criteria are met}

\STATE \textbf{Step C: Image classification}
    \STATE Perform the prediction with the final classifier using the feature vectors \( \{F_1, F_2, \dots, F_M\} \).

\end{algorithmic}
\end{algorithm}

\subsubsection{Step A: Seed Selection}

This step involves two key components: the CLIP model, serving as the Image/Text encoder, and a pre-trained feature extractor (FE). The primary goal of this step is to construct the $SEED$ set of images with pseudo-labels by leveraging CLIP's capability to encode images and text into a shared latent space. This allows us to compare class names (labels) and images by evaluating the similarity of their embeddings. A higher similarity score between the embedding of an image $i$ and a label $l$ indicates a higher likelihood that $l$ is the correct label for $i$. Following the default CLIP preprocess for text, we adopt the descriptive phrase format, such as “{This is a photo of a $l$}", for encoding labels. In this step, we consider the cosine similarity (the standard measurement used by CLIP) between the text and image embeddings, defined as:

\begin{equation}
\begin{aligned}
\text{cos}(I, E) &= \frac{I \cdot E}{\|I\| \|E\|}.
\end{aligned}
\label{eq:cos}
\end{equation}

Where $I$ represents the embedding of an image and $E$ represents the embedding of a text description. In our approach, we first use CLIP\textsubscript{text} to encode each label into a textual embedding and CLIP\textsubscript{image} to encode the dataset images. The cosine similarity between these embeddings is then calculated for all label-image pairs. Based on the resulting similarity scores, we select the 100 images most similar to each label \( l \in \mathcal{L} \), forming a set \( \mathcal{B}(l) \) for each label. This process is referred to as the \textbf{default selection}.

We further enhance this selection process by introducing a scoring mechanism to estimate the likelihood of an image belonging to a given label \( l \). For each image \( x \in \mathcal{B}(l) \), we perform the following steps:
\begin{enumerate}
    \item Retrieve the top \( k \) most similar images in the dataset to the embedding \( \mathcal{I}(x) \), where \( \mathcal{I}(x) \) denotes the image embedding of \( x \). Denote these retrieved images as \( r_j(x) \) for \( j = 1, \dots, k \).
    
    \item For each retrieved image \( r_j(x) \), predict the similarity with the labels using CLIP (as described previously and in Section~\ref{sec:rel_works}).
    
    \item Compute a similarity score \( S(x) \) for image \( x \) as follows:
    \begin{equation}
        S(x) = \frac{1}{k} \sum_{j=1}^{k} \cos(\mathcal{I}(r_j(x)), \mathcal{T}(l_i)),
        \label{eq:score_function}
    \end{equation}
    where \( \mathcal{C}_j(l) \) denotes the cosine similarity between label \( l \) and the retrieved image \( r_j(x) \).
\end{enumerate}

This score reflects how consistently the retrieved images are associated with the target label \( l \), indicating the reliability of image \( i \)'s association with that label. This approach to constructing the score was developed through a series of empirical experiments. The intuition is that if an image is a strong candidate for a class, its nearest neighbors in the feature space should also be consistently associated with that same class label. This mechanism, based on \emph{neighborhood consensus}, provides a more robust confidence measure than relying on the candidate image's similarity alone.
Using the score, we construct a ranking 
\( \mathcal{R}(l) \) for each known label \( l \in \mathcal{L} \), ordering the set of images from the most to the least related to the label $l$. This process is referred to as the \textbf{improved selection}.

Each label's top \( k \) images are selected and assigned pseudo-labels based on their corresponding rankings, forming the $SEED$ set.

\subsubsection{Step B: Classifier Training}

The $SEED$ set is used to initialize the training of a lightweight classifier that consists of just a single layer with a linear activation function followed by a softmax function. To enhance the training process of the classifier in Step B, we employ a pre-trained feature extractor to generate representative feature vectors for all dataset images. This approach leverages transfer learning, allowing us to train a robust classifier even with relatively few images. Each sample used in this step (including the $SEED$ set) is represented as a pair $(X,y)$, where \(X \in \mathbb{R}^d\) represents the d-dimensional feature vector extracted by the pre-trained model capturing rich and discriminative characteristics of a given image $i$; while $y \in L$ corresponds to the pseudo-label assigned to $i$. The classifier is configured with an output size that corresponds to the number of classes. Training begins with an initial phase lasting $I_{epochs}$, using the $SEED$ set selected during Step A.
After completing the initial training, we transition to a self-learning cycle. In each cycle, we iterate through the ranking 
$R(l)$ for each label $l_i \in L$ to select the next top-\( k \) similar images of each known label. The current classifier evaluates these $k \times |L|$ images, and their logits are used to assign a predicted label $\hat{y} = \arg\max(\text{logits})$. If \(\hat{y}\) matches the label $l$ corresponding to the ranking from which the image was selected, the image is confidently pseudo-labeled and is added to the tuning set $\mathcal{D}_{\text{tune}}$. The classifier is then fine-tuned using $\mathcal{D}_{\text{tune}}$ for $R_{epochs}$. For each new iteration cycle, the tuning set is reset ($D_{\text{tune}} \leftarrow \emptyset$). The self-learning process continues until one of the stopping criteria is met:

\begin{equation}
\begin{aligned}
\text{Stop when } & \ \text{Loss}_{new} \geq \text{Loss}_{prev} \\
& \text{or } \text{Loss}_{epoch} \geq \text{Loss}_{limit} \\
& \text{or } \text{Cycle count} = \text{$Max Cycles$}.
\end{aligned}
\label{eq:loss}
\end{equation}

$Loss_{prev}$ represents the first loss value of the last training step of the self-learning cycle, and $Loss_{new}$ is the first loss value of the current training step. This first criterion ($\text{Loss}_{new} \geq \text{Loss}_{prev}$) is used to prevent overfitting by comparing the last loss of the previous cycle with the first loss of the current cycle. $Loss_{epoch}$ represents the loss at the current iteration of the self-learning cycle, while $Loss_{limit}$ is a hyperparameter that defines a threshold above which the training loss is considered too high to continue improving the model. Therefore, if  $\text{Loss}_{epoch} \geq \text{Loss}_{limit}$, it indicates that the training process has reached its predefined threshold to prevent overfitting. $MaxCycles$ is another hyperparameter that dictates the maximum number of training cycles in the pipeline. These criteria promote convergence by monitoring loss stability or reaching a predefined loss or cycle limit.
Upon completing the self-learning cycle, we obtain a trained classifier.

\subsubsection{Step C: Image Classification}

In this step, we predict the class of each image in the dataset by applying the final classifier to the extracted features. Initially, the feature extractor $FE$ is used to obtain features for all images via \( FE(x) \), resulting in a set of feature vectors \( \{F_1, F_2, \dots, F_M\} \). These feature vectors are then processed by the final classifier, which was trained in step B. The classifier outputs the probability distribution over the classes for each image. The predicted class for each image is then determined as $\hat{y} = \arg\max(\text{logits})$.

\section{Experiments} \label{sec:experiments}

This section describes the experiments carried out to evaluate the performance of our approach on different and varied datasets. Firstly, we introduce the datasets used in our experiments. Next, we outline the methodology employed in our experiments. Lastly, we present our analyses based on the experimental results.

\subsection{Datasets}
\label{datasets}

To evaluate our method, we performed experiments on 10 widely adopted datasets: ImageNet \cite{deng:09}, Stanford Cars \cite{KrauseStarkDengFei:13}, Caltech-101 \cite{fei2006one}, Flowers \cite{nilsback2008automated}, CIFAR-10 \cite{krizhevsky:09}, CIFAR-100 \cite{krizhevsky:09}, Aircraft \cite{maji13fine-grained}, Caltech-256 \cite{griffin2007caltech}, Textures \cite{cimpoi2014describing}, and Food-101 \cite{bossard2014food}. All these datasets are colorful, with images depicting different contexts focusing on different categories and with different characteristics. Some encompass fine-grained classes, while others distinguish each image category more. These varied characteristics provide a more generalized assessment of the approach's performance. Table \ref{tab:datasetTable} shows essential information about these datasets. Since our goal is to evaluate the approach directly on data for inference, particularly with large amounts of unlabeled data, we handled the datasets in different ways. For datasets with a sufficiently large evaluation or test set, such as ImageNet (evaluation set) and CIFAR (test set), we used only these splits. In contrast, the Caltech-101 and Caltech-256 datasets do not provide predefined splits, so we used their full datasets by default. For the remaining datasets, we merged all available splits to ensure a sufficient number of samples for our experiments.


\begin{table}[!ht]
\centering
\caption{Datasets information}
\label{tab:datasetTable}
\begin{tabular}{|l|c|c|c|}
\hline
\textbf{Dataset}                                                     & \textbf{Instances}    & \textbf{Classes}      & \multicolumn{1}{l|}{\textbf{\begin{tabular}[c]{@{}l@{}}Avg Instances $\pm$ Std\\ per Class\end{tabular}}} \\ \hline
\textbf{ImageNet \cite{deng:09}}                                                    & 50000                      & 1000                      & 50 $\pm$ 0                                                                                                          \\ \hline
\textbf{Stanford Cars\cite{KrauseStarkDengFei:13}} & 16185                 & 196                   & 82.58 $\pm$ 8.69                                                                                             \\ \hline
\textbf{Caltech-101 \cite{fei2006one}}                                                 &  9144                     & 101                      &  89.65 $\pm$ 123.65                                                                                                      \\ \hline
\textbf{Flowers \cite{nilsback2008automated}}                                                     &  8189                      &   102                    & 80.28 $\pm$ 44.28                                                                                                           \\ \hline
\textbf{CIFAR-10 \cite{krizhevsky:09}}               & 10000                 & 10                    & 1000 $\pm$ 0                                                                                              \\ \hline
\textbf{CIFAR-100 \cite{krizhevsky:09}}              & 10000                 & 100                   & 100 $\pm$ 0                                                                                               \\ \hline
\textbf{Aircraft \cite{maji13fine-grained}}         & 10000                 & 100                   & 100 $\pm$ 0                                                                                               \\ \hline
\textbf{Caltech-256 \cite{griffin2007caltech}}                                                 & 30607 & 256 & 119.09 $\pm$ 85.85                                                                                      \\ \hline
\textbf{Textures \cite{cimpoi2014describing}}                                                    & 5640 & 47 & 120 $\pm$ 0                                                                                     \\ \hline
\textbf{Food-101 \cite{bossard2014food}}                                                        & 25250 & 101 & 250  $\pm$ 0                                                                                      \\ \hline
\end{tabular}

\end{table}

\subsection{Methodology}
\label{methodo}

In the initial stage of our experimental analysis, we evaluate the accuracy of the developed seed selection approach (Step A, Section \ref{sec:approach}) to understand its behavior and performance as the parameter $k$ varies. Here, $k$ represents the number of images selected and considered as belonging to a given class without prior knowledge of their true labels. We tested seven different values of $k$ and compared the two proposed methodologies: the default selection and the improved selection. This analysis aims to identify how many of the samples chosen by the seed generation process actually belong to the assigned classes.

To evaluate the zero-shot image classification, we followed the same methodology adopted in \cite{mirza2023lafter,pratt2023does}, using accuracy as the evaluation metric and compared the performance of our approach with the performance of the standard CLIP\cite{radford2021learning} and a state-of-the-art approach called AdaptCLIPZS \cite{saha2024improved}. Because our approach is entirely zero-shot, we considered only the zero-shot variants of \cite{saha2024improved} for comparison, leaving out the fine-tuning approach. To ensure methodological consistency, we adopted the descriptions provided by the authors and employed their algorithm and the same LLM model (gpt-4-0613) to generate descriptions for the datasets they did not use in their work.

We use three CLIP backbones in the evaluation: ViT-B/16, ViT-B/32, and ViT-L/14. We evaluated our approach in two different ways: using only the $SEED$ for training (we call it $Seed$ in the results), and using the complete approach with the self-learning cycle ($Complete$). Different runs have an observed standard deviation low, ranging from 0.1\% to 0.2\% (thus being omitted from the results table), indicating the stability of the method. 

Our approach can adopt different models for image/text encoders, feature extractors, and image classifiers. For the feature extractor, based on recent results \cite{iceis24erick,todescato2024investigating} and empirical experiments, we choose ViT-G-14 \cite{rw2019timm} pre-trained with the LAION-2B English subset of LAION-5B using OpenCLIP \cite{ilharco_gabriel_2021_5143773}. The weights of this model are publicly available \footnote{Can be accessed through \url{https://github.com/huggingface/pytorch-image-models}}. We adopted a lightweight model with a linear activation function and a softmax as the image classifier. 

For each dataset, the images went through a homogeneous pre-processing. The images are resized and center-cropped to size 224*224, the input size of CLIP, and the feature extractor. We adopted the Adam Optimizer with a learning rate of $0.001$ and cross-entropy loss to train the classifier. The values for the approach's hyperparameters were $I_{epochs}=100$ and $R_{epochs}=20$, and $k=5$, defined by empirical experiments. This means that the initial training was performed for 100 epochs, and in each iteration of the self-learning cycle, the tuning was performed for 20 epochs, and we selected five images of each class using the ranking for the $SEED$ and kept the same number of images for analyses in each cycle step. This configuration is used in all datasets except ImageNet and Caltech-256, where, due to its high number of classes, we reduced the learning rate to $0.0001$ and increased the number of samples for the $SEED$ setting $k=16$.
The $ \text{Loss}_\text{limit} $ in the self-learning cycle equals $0.1$. We define $MaxCycles=20$ and thus we limit the size of \( \mathcal{B}(l_i) \) and \( \mathcal{R}(l_i) \) (i.e., the ranking for each class) to 100th position.

Our code was implemented in Python, using mainly the PyTorch library\footnote{\url{https://pytorch.org/}} and is available at Anonymous Github \footnote{\url{https://github.com/mvtodescato/ZS_ImageClassification}}.

\subsection{Results}

\begin{figure}[!ht]
\centering
\includegraphics[width=0.48\textwidth]{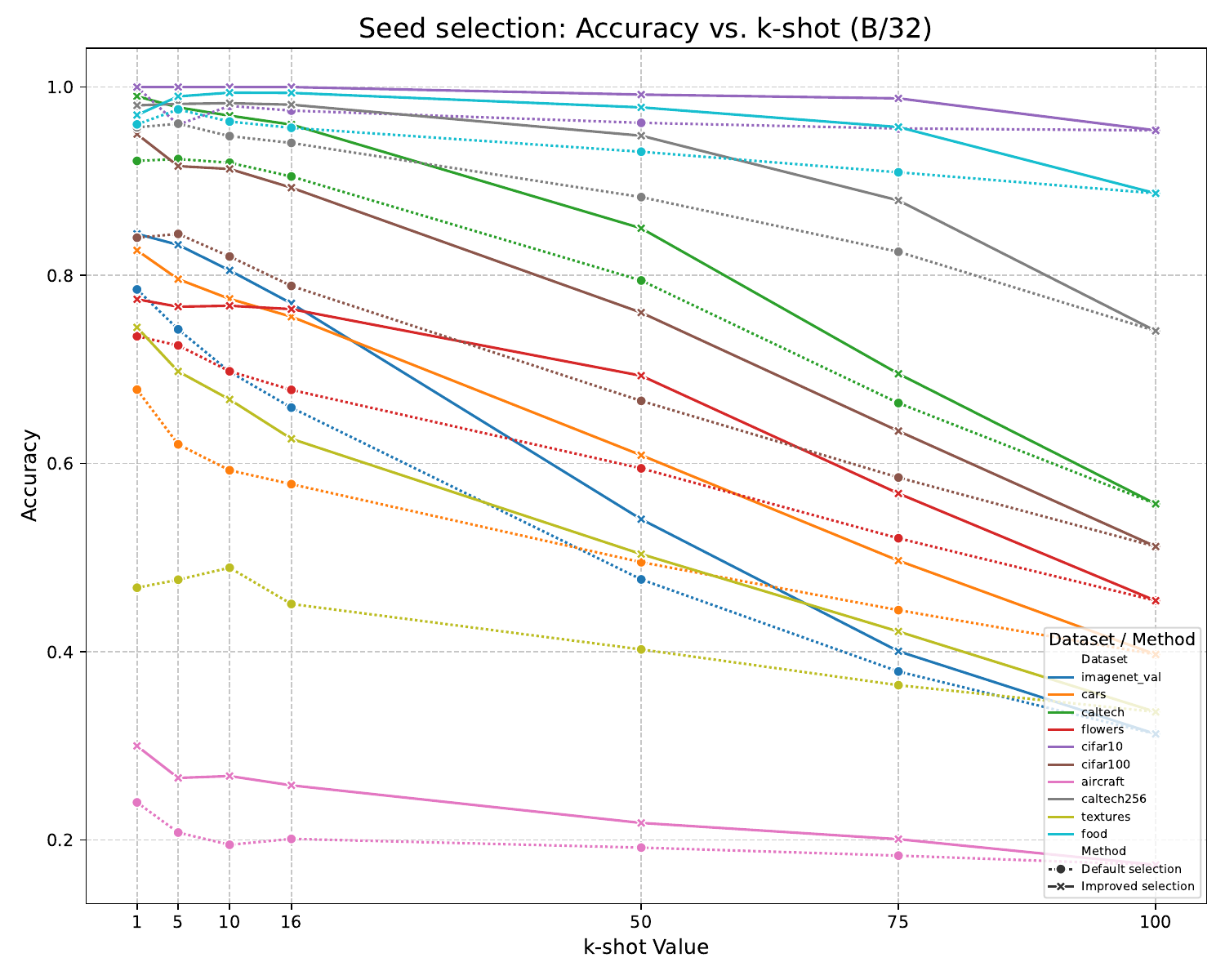}
\caption{Accuracy of the sample selection (default and improved versions) across the datasets using CLIP B/32 backbone}
\label{fig:b32}
\end{figure}

\begin{figure}[!ht]
\centering
\includegraphics[width=0.48\textwidth]{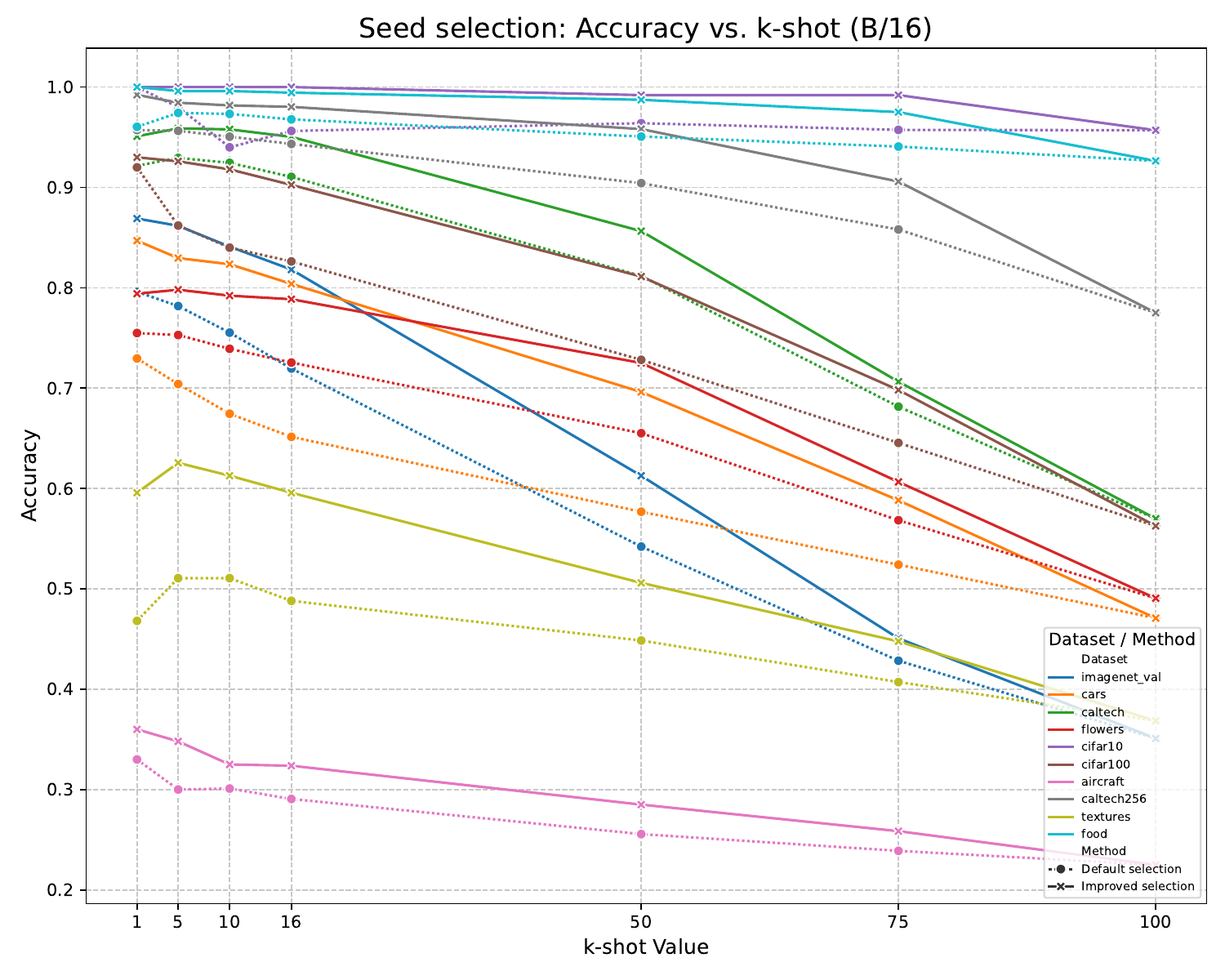}
\caption{Accuracy of the sample selection (default and improved versions) across the datasets using CLIP B/16 backbone}
\label{fig:b16}
\end{figure}

\begin{figure}[!ht]
\centering
\includegraphics[width=0.48\textwidth]{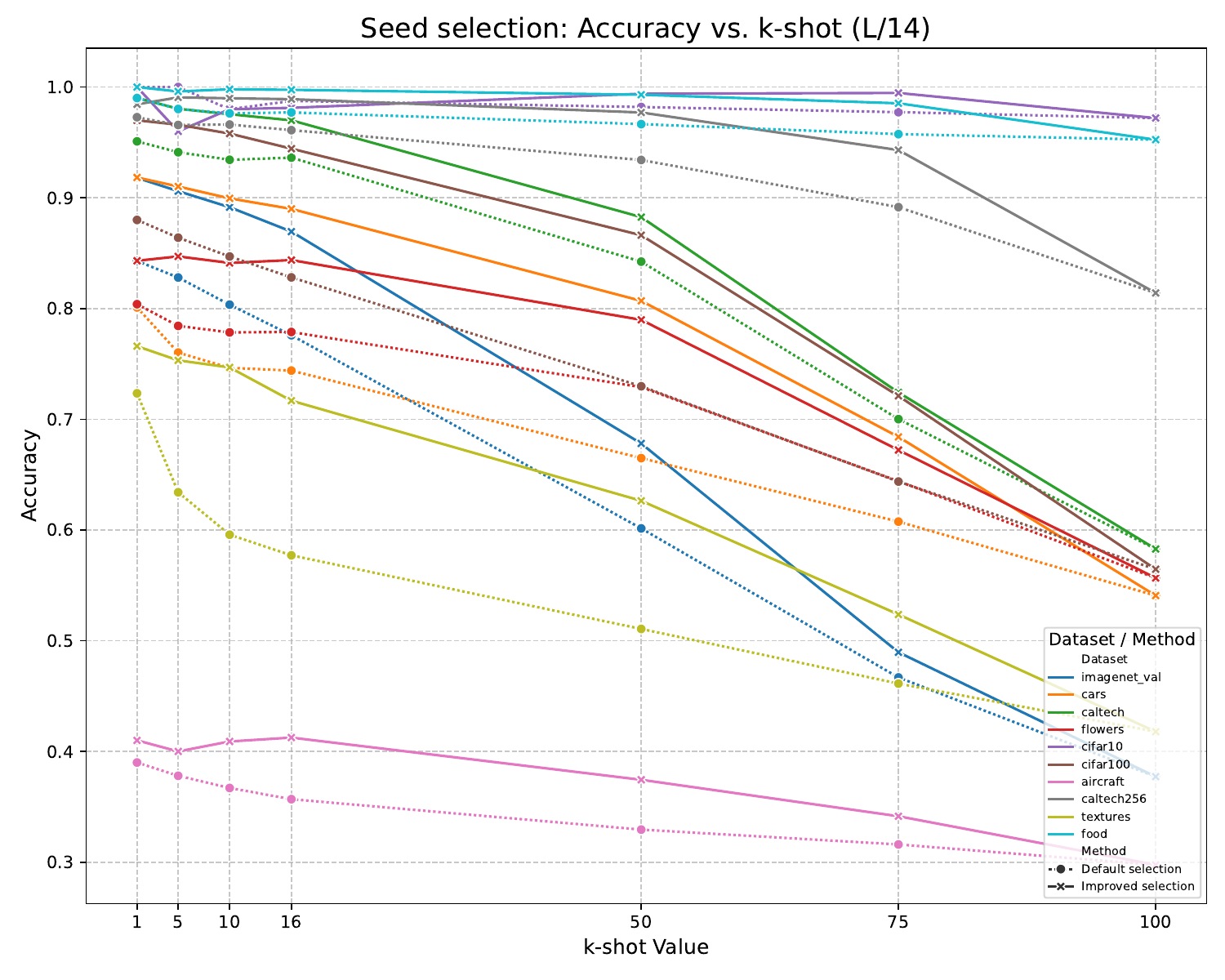}
\caption{Accuracy of the sample selection (default and improved versions) across the datasets using CLIP L/14 backbone}
\label{fig:l14}
\end{figure}

In this section, we discuss the results of our experiments. In our first set of analyses, we compare the default and improved seed selection. In this experiment, we identify how many samples selected by the seed generation process are actually from the classes that were assigned. Figures \ref{fig:b32}, \ref{fig:b16}, and \ref{fig:l14} show the accuracy of the sample selection with seven variations of the value of $k$ within the set of 100 images initially selected in the three CLIP models used. 

\begin{table*}[!ht]
\centering
\caption{Accuracy of each approach in each dataset.The highest accuracy is shown in bold, while the second best is underlined..}
\label{tab:resultsTable}
\resizebox{0.9\textwidth}{!}{
\begin{tabular}{|lccccccccccc|}
\hline
\multicolumn{1}{|c|}{\multirow{2}{*}{\textbf{}}} & \multicolumn{10}{c|}{\textbf{Dataset}}                                                                                                                                                                                                                                                                                                                                                            & \multirow{2}{*}{\textbf{Average}} \\
\multicolumn{1}{|c|}{}                           & \textbf{Imagenet}                   & \textbf{Cars}                       & \multicolumn{1}{l}{\textbf{Caltech-101}} & \multicolumn{1}{l}{\textbf{Flowers}} & \textbf{CIFAR-10}                    & \textbf{CIFAR-100}                   & \textbf{Aircraft}                   & \textbf{Caltech-256}                & \multicolumn{1}{l}{\textbf{Textures}} & \multicolumn{1}{l|}{\textbf{Food}}  &                                   \\ \hline
\textbf{B/32}                                    &                                     &                                     & \multicolumn{1}{l}{}                     & \multicolumn{1}{l}{}                 &                                     &                                     &                                     &                                     & \multicolumn{1}{l}{}                  & \multicolumn{1}{l}{}                &                                   \\ \hline
\multicolumn{1}{|l|}{CLIP}                       & \multicolumn{1}{c|}{61.15}          & \multicolumn{1}{c|}{57.62}          & \multicolumn{1}{c|}{81.45}               & \multicolumn{1}{c|}{63.99}           & \multicolumn{1}{c|}{87.65}          & \multicolumn{1}{c|}{59.54}          & \multicolumn{1}{c|}{17.48}          & \multicolumn{1}{c|}{84.04}          & \multicolumn{1}{c|}{43.47}            & \multicolumn{1}{c|}{{\ul 86.51}}    & 64.29                             \\ \hline
\multicolumn{1}{|l|}{AdaptCLIPZS (D)}            & \multicolumn{1}{c|}{61.32}          & \multicolumn{1}{c|}{57.75}          & \multicolumn{1}{c|}{82.05}               & \multicolumn{1}{c|}{66.39}           & \multicolumn{1}{c|}{87.81}          & \multicolumn{1}{c|}{62.27}          & \multicolumn{1}{c|}{18.64}          & \multicolumn{1}{c|}{82.04}          & \multicolumn{1}{c|}{43.73}            & \multicolumn{1}{c|}{82.99}          & 64.49                             \\ \hline
\multicolumn{1}{|l|}{AdaptCLIPZS (A)}            & \multicolumn{1}{c|}{63.11}          & \multicolumn{1}{c|}{57.73}          & \multicolumn{1}{c|}{85.68}               & \multicolumn{1}{c|}{68.08}           & \multicolumn{1}{c|}{87.40}          & \multicolumn{1}{c|}{61.52}          & \multicolumn{1}{c|}{19.19}          & \multicolumn{1}{c|}{85.50}          & \multicolumn{1}{c|}{46.82}            & \multicolumn{1}{c|}{82.94}          & 65.80                             \\ \hline
\multicolumn{1}{|l|}{Ours (Seed)}                & \multicolumn{1}{c|}{\textbf{71.09}} & \multicolumn{1}{c|}{{\ul 78.39}}    & \multicolumn{1}{c|}{{\ul 87.11}}         & \multicolumn{1}{c|}{{\ul 70.47}}     & \multicolumn{1}{c|}{{\ul 93.34}}    & \multicolumn{1}{c|}{{\ul 77.30}}    & \multicolumn{1}{c|}{{\ul 23.38}}    & \multicolumn{1}{c|}{\textbf{90.46}} & \multicolumn{1}{c|}{{\ul 50.08}}      & \multicolumn{1}{c|}{86.48}          & {\ul 72.81}                       \\ \hline
\multicolumn{1}{|l|}{Ours (Complete)}            & \multicolumn{1}{c|}{{\ul 68.29}}    & \multicolumn{1}{c|}{\textbf{79.63}} & \multicolumn{1}{c|}{\textbf{87.12}}      & \multicolumn{1}{c|}{\textbf{75.16}}  & \multicolumn{1}{c|}{\textbf{95.07}} & \multicolumn{1}{c|}{\textbf{80.14}} & \multicolumn{1}{c|}{\textbf{24.44}} & \multicolumn{1}{c|}{{\ul 89.75}}    & \multicolumn{1}{c|}{\textbf{51.96}}   & \multicolumn{1}{c|}{\textbf{87.95}} & \textbf{73.95}                    \\ \hline
\textbf{B/16}                                    &                                     & \multicolumn{1}{l}{}                & \multicolumn{1}{l}{}                     & \multicolumn{1}{l}{}                 & \multicolumn{1}{l}{\textbf{}}       & \multicolumn{1}{l}{\textbf{}}       & \multicolumn{1}{l}{}                & \multicolumn{1}{l}{}                & \multicolumn{1}{l}{}                  & \multicolumn{1}{l}{}                & \multicolumn{1}{l|}{}             \\ \hline
\multicolumn{1}{|l|}{CLIP}                       & \multicolumn{1}{c|}{66.03}          & \multicolumn{1}{c|}{63.73}          & \multicolumn{1}{c|}{81.22}               & \multicolumn{1}{c|}{65.16}           & \multicolumn{1}{c|}{89.35}          & \multicolumn{1}{c|}{63.29}          & \multicolumn{1}{c|}{22.40}          & \multicolumn{1}{c|}{81.87}          & \multicolumn{1}{c|}{42.83}            & \multicolumn{1}{c|}{80.59}          & 65.65                             \\ \hline
\multicolumn{1}{|l|}{AdaptCLIPZS (D)}            & \multicolumn{1}{c|}{66.24}          & \multicolumn{1}{c|}{63.88}          & \multicolumn{1}{c|}{81.58}               & \multicolumn{1}{c|}{70.94}           & \multicolumn{1}{c|}{90.10}          & \multicolumn{1}{c|}{66.86}          & \multicolumn{1}{c|}{24.67}          & \multicolumn{1}{c|}{84.71}          & \multicolumn{1}{c|}{44.05}            & \multicolumn{1}{c|}{87.98}          & 68.10                             \\ \hline
\multicolumn{1}{|l|}{AdaptCLIPZS (A)}            & \multicolumn{1}{c|}{68.74}          & \multicolumn{1}{c|}{63.62}          & \multicolumn{1}{c|}{\textbf{87.33}}      & \multicolumn{1}{c|}{72.56}           & \multicolumn{1}{c|}{90.56}          & \multicolumn{1}{c|}{66.51}          & \multicolumn{1}{c|}{24.26}          & \multicolumn{1}{c|}{87.56}          & \multicolumn{1}{c|}{\textbf{48.02}}   & \multicolumn{1}{c|}{\textbf{88.50}} & 69.77                             \\ \hline
\multicolumn{1}{|l|}{Ours (Seed)}                & \multicolumn{1}{c|}{\textbf{72.52}} & \multicolumn{1}{c|}{{\ul 77.98}}    & \multicolumn{1}{c|}{82.89}               & \multicolumn{1}{c|}{{\ul 75.43}}     & \multicolumn{1}{c|}{{\ul 93.79}}    & \multicolumn{1}{c|}{{\ul 77.60}}    & \multicolumn{1}{c|}{{\ul 25.36}}    & \multicolumn{1}{c|}{{\ul 90.83}}    & \multicolumn{1}{c|}{44.49}            & \multicolumn{1}{c|}{86.70}          & {\ul 72.76}                       \\ \hline
\multicolumn{1}{|l|}{Ours (Complete)}            & \multicolumn{1}{c|}{{\ul 70.61}}    & \multicolumn{1}{c|}{\textbf{79.60}} & \multicolumn{1}{c|}{{\ul 82.93}}         & \multicolumn{1}{c|}{\textbf{77.82}}  & \multicolumn{1}{c|}{\textbf{95.14}} & \multicolumn{1}{c|}{\textbf{80.81}} & \multicolumn{1}{c|}{\textbf{28.05}} & \multicolumn{1}{c|}{\textbf{91.05}} & \multicolumn{1}{c|}{{\ul 46.34}}      & \multicolumn{1}{c|}{{\ul 87.88}}    & \textbf{74.02}                    \\ \hline
\textbf{L/14}                                    &                                     &                                     &                                          &                                      &                                     &                                     &                                     &                                     &                                       &                                     &                                   \\ \hline
\multicolumn{1}{|l|}{CLIP}                       & \multicolumn{1}{c|}{72.58}          & \multicolumn{1}{c|}{76.38}          & \multicolumn{1}{c|}{85.78}               & \multicolumn{1}{c|}{74.27}           & \multicolumn{1}{c|}{95.56}          & \multicolumn{1}{c|}{73.24}          & \multicolumn{1}{c|}{30.53}          & \multicolumn{1}{c|}{87.10}          & \multicolumn{1}{c|}{51.37}            & \multicolumn{1}{c|}{91.66}          & 73.85                             \\ \hline
\multicolumn{1}{|l|}{AdaptCLIPZS (D)}            & \multicolumn{1}{c|}{73.08}          & \multicolumn{1}{c|}{76.13}          & \multicolumn{1}{c|}{84.84}               & \multicolumn{1}{c|}{78.26}           & \multicolumn{1}{c|}{95.20}          & \multicolumn{1}{c|}{76.60}          & \multicolumn{1}{c|}{{\ul 32.82}}    & \multicolumn{1}{c|}{87.31}          & \multicolumn{1}{c|}{54.49}            & \multicolumn{1}{c|}{92.97}          & 75.17                             \\ \hline
\multicolumn{1}{|l|}{AdaptCLIPZS (A)}            & \multicolumn{1}{c|}{\textbf{75.69}} & \multicolumn{1}{c|}{76.17}          & \multicolumn{1}{c|}{\textbf{89.65}}      & \multicolumn{1}{c|}{78.50}           & \multicolumn{1}{c|}{\textbf{95.35}} & \multicolumn{1}{c|}{76.80}          & \multicolumn{1}{c|}{\textbf{34.18}} & \multicolumn{1}{c|}{91.60}          & \multicolumn{1}{c|}{\textbf{57.72}}   & \multicolumn{1}{c|}{\textbf{93.24}} & {\ul 76.89}                       \\ \hline
\multicolumn{1}{|l|}{Ours (Seed)}                & \multicolumn{1}{c|}{{\ul 74.55}}    & \multicolumn{1}{c|}{{\ul 84.90}}    & \multicolumn{1}{c|}{{\ul 87.82}}         & \multicolumn{1}{c|}{{\ul 78.51}}     & \multicolumn{1}{c|}{93.52}          & \multicolumn{1}{c|}{{\ul 79.50}}    & \multicolumn{1}{c|}{29.72}          & \multicolumn{1}{c|}{{\ul 91.85}}    & \multicolumn{1}{c|}{54.97}            & \multicolumn{1}{c|}{86.08}          & 76.14                             \\ \hline
\multicolumn{1}{|l|}{Ours (Complete)}            & \multicolumn{1}{c|}{71.47}          & \multicolumn{1}{c|}{\textbf{86.19}} & \multicolumn{1}{c|}{87.29}               & \multicolumn{1}{c|}{\textbf{80.61}}  & \multicolumn{1}{c|}{{\ul 95.13}}    & \multicolumn{1}{c|}{\textbf{83.29}} & \multicolumn{1}{c|}{29.53}          & \multicolumn{1}{c|}{\textbf{92.10}} & \multicolumn{1}{c|}{{\ul 56.56}}      & \multicolumn{1}{c|}{{\ul 87.56}}    & \textbf{76.97}                    \\ \hline
\end{tabular}
}

\end{table*}

First, it is necessary to highlight the potential of sample selection using the default selection. In the configuration used for the experiments ($k=5$), the default selection using the least effective backbone (B/32) achieves an accuracy above 80\% in 5 of the 10 datasets and above 60\% in 8 of the 10 datasets, with an average of 64.77\% accuracy in the selection. Under these same conditions, the improved selection presents a significant improvement compared to the default, presenting an average of 82.25\% accuracy, representing an improvement of 17.48 percentage points (pp) in seed selection. These numbers highlight that despite being the worst backbone, the approach selects the seed with reasonable accuracy to start the training process.  With the improvement in the backbones, seed selection becomes even more precise. With $k=5$ and the B/16 backbone, the improved selection accuracy reaches an average of 83.28\%, and with the L/14 backbone, an average of 87.08\%, even though the aircraft dataset has only 40\% accuracy under these conditions. From the charts, the improvement of the improved selection compared to the default becomes smaller as the backbones improve, but is still significant.
Another important thing to note is that in most datasets, there is a loss of accuracy when $k=10$ compared to $k=5$ and an even greater loss when $k=16$ and $k=50$. This loss justifies using $k=5$, since $k=1$ does not provide enough information for the classifier. Then the choice of $k=5$ focuses on having good accuracy in the initial seed selection since errors can increase a confirmation bias, damaging the self-training cycle. The ImageNet dataset is an exception because even with a high number of samples selected using $k=16$ (i.e., 16000 images), the accuracy does not decrease that much, and according to empirical tests, improving the results of the approach.

Regarding the zero-shot image classification, the experimental results are summarized in Table \ref{tab:resultsTable}. We compare our method against the baseline CLIP and with the state-of-the-art method AdaptCLIPZS. 
In the table, \textbf{CLIP} refers to the baseline CLIP tested with the default prompt “photo of a [\textit{class name}]". \textbf{AdaptCLIPZS (D)} refers to CLIP tested with the prompt “a photo of a [\textit{class name}] [\textit{domain}]". \textbf{AdaptCLIPZS (A)} is evaluating CLIP with the prompt “a photo of a [\textit{class name}] [\textit{domain}] with [\textit{characteristic}]" using the descriptions obtained from the LLM model.
Our results indicate that our approach achieves superior performance on average for all the backbones. First, analyzing the results obtained with the B/32 backbone, we can see that our approach outperforms CLIP and AdaptCLIPZS in all datasets, with the complete approach (with the self-training cycle) improving the results of using only the seed in 8 of the 10 datasets. Our complete approach improves 9.66 pp compared to CLIP and 8.15 pp compared to AdaptCLIPZS (A) in this configuration. With the B/16 backbone, the result is similar, but with AdaptCLIPZS (A) being superior in the Textures, Food and Caltech-101 datasets. In this configuration, only one dataset achieved better results using just the initial seed compared to the full self-learning cycle.

Using the L/14 backbone, the gap between our method and vanilla CLIP becomes small. Our approach and AdaptCLIPZS share the best results, with our approach having a slight advantage, delivering the highest mean accuracy overall, albeit by a narrow margin. On average, we surpass vanilla CLIP by $\approx$ 3 percentage points (pp) in this configuration, and by approximately 10 pp on Cars and CIFAR-100. Although the complete self-learning cycle lowers accuracy on a few datasets, it still presents a significant improvement in several datasets and in the overall result.

We can see that the improvement in the backbone reduced the gap between our two approaches and invalidated the use of the training cycle in some cases. In these cases, there is overfitting and a confirmation bias in the training cycle. In the ImageNet dataset, for example, our approach loses much potential with the training cycle, resulting in a sudden loss of accuracy. The small number of images per class and the high number of classes in ImageNet tend to be the reason for this drop, a characteristic that also occurs in the Caltech datasets. We hypothesize this is due to error accumulation; with 1000 classes, even a small error rate in pseudo-labeling can introduce significant noise into the training set, which the classifier then overfits. Furthermore, the performance drop in the self-learning cycle is also influenced by the model's reliance on semantics, a challenge particularly evident in fine-grained datasets like Aircraft. However, in datasets such as CIFAR-100, for example, we have an improvement of almost 4 pp in accuracy with the complete approach compared with only the seed. This shows that the complete approach significantly improves performance in cases where the semantics of the classes are distinct and we have a reasonable amount of samples per class.

\section{Conclusions} \label{sec:conclusion}

This paper introduces a novel approach for zero-shot image classification based on a self-learning cycle. By combining the zero-shot capabilities of CLIP with features from a pre-trained transformer, we propose a modular and efficient framework that eliminates the need for labeled data. Our method uses CLIP to select samples and employs a self-learning cycle to iteratively refine pseudo-labels, relying only on a set of class names. Extensive experiments across ten different datasets demonstrate that our approach outperforms the baseline CLIP in every backbone, being superior to the state-of-the-art approach in two of them and equally effective in the last one. Notably, on average, our approach surpasses the CLIP baseline by $\approx$3 pp to $\approx$10 pp depending on the backbone used. These findings highlight the effectiveness of our approach and the benefits of leveraging generalized feature extraction for zero-shot image classification. Furthermore, the results of seed selection accuracy using CLIP demonstrate the effectiveness of this approach and the possibility of using this strategy in other tasks. In addition, besides its strong performance, our approach is modular and adaptable. The framework supports the integration of different backbone models.  In future works, we plan to investigate the use of our approach in a few-shot setting, evaluating the seed selection process with labeled samples.
Despite its strengths, our approach is still dependent on the performance of CLIP and the semantic condition of the dataset (whether they are distinct or common words). Future work will focus on refining these aspects to improve robustness and adaptability.

\bibliographystyle{IEEEtran}
\bibliography{biblio}

\end{document}